\documentclass[twoside,11pt]{article}

\usepackage{jmlr2e}
\usepackage{graphicx}
\usepackage{float}
\usepackage{listings}
\usepackage{mathtools}
\usepackage{pgfplots}
\usepackage[super]{nth}

\pgfkeys{/pgf/declare function={arcsinh(\x) = ln(\x + sqrt(\x^2+1));}}
\pgfplotsset{compat=1.14}

\ShortHeadings{m-ar-K-FastICA: Reliable Feature Extraction in scikit-learn}{Luca Parisi, \emph{PhD, MBA Candidate}}
\firstpageno{1}

\begin{document}

\title{m-ar-K-Fast Independent Component Analysis}

\author{\name Luca Parisi \email luca.parisi@ieee.org \\
       \addr Coventry, United Kingdom \\
       PhD in Machine Learning for Clinical Decision Support Systems \\ 
       MBA Candidate with Artificial Intelligence Specialism
       }

\editor{}

\maketitle


\begin{abstract}
This study presents the m-arcsinh Kernel ('m-ar-K') Fast Independent Component Analysis ('FastICA') method ('m-ar-K-FastICA') for feature extraction. The kernel trick has enabled dimensionality reduction techniques to capture a higher extent of non-linearity in the data; however, reproducible, open-source kernels to aid with feature extraction are still limited and may not be reliable when projecting features from entropic data. The m-ar-K function, freely available in Python and compatible with its open-source library 'scikit-learn', is hereby coupled with FastICA to achieve more reliable feature extraction in presence of a high extent of randomness in the data, reducing the need for pre-whitening. Different classification tasks were considered, as related to five (N = 5) open access datasets of various degrees of information entropy, available from scikit-learn and the University California Irvine (UCI) Machine Learning repository. Experimental results demonstrate improvements in the classification performance brought by the proposed feature extraction. The novel m-ar-K-FastICA dimensionality reduction approach is compared to the 'FastICA' gold standard method, supporting its higher reliability and computational efficiency, regardless of the underlying uncertainty in the data.

\end{abstract}

\vskip 0.1in


\newpage


\vskip 0.3in

\section{Introduction}

\vskip 0.15in

Despite theoretical advances in the Independent Component Analysis (ICA) feature extraction method, suitable for dimensionality reduction in presence of non-Gaussian/non-normal data distributions, such as the FastICA~\citep{hyvarinen2000independent}~\citep{javidi2011fast}, the Kernel ICA (KICA)~\citep{bach2002kernel}, the Fast Kernel-based ICA (FastKICA)~\citep{shen2009fast}, and the Adaptive Mixture ICA (AMICA)~\citep{hsu2018modeling}, usable, explainable, reproducible and replicable variations of the ICA method that improve feature extraction are still limited. In fact, the open-source Python library named 'scikit-learn'~\citep{scikit-learn} leverages the FastICA algorithm as the only ICA-related variation available, which uses the ~\emph{logcosh} as the G function to approximate to neg-entropy~\citep{nguyen2018novel}.
\\

However, such variants of ICA, even with appropriate data pre-whitening~\citep{james2004independent}~\citep{hsu2015real}, may not be reliable when performing dimensionality reduction of entropic data~\citep{mckeown2003independent}~\citep{li2010complex}, characterised by a high extent of randomness or uncertainty. When applied prior to classification, they can lead to slow or lack of convergence~\citep{vert2006consistency}~\citep{jacot2018neural}, due to trapping at local \emph{minima}~\citep{parisi2014exploiting}~\citep{parisi2016preliminary}~\citep{parisi2017importance}~\citep{parisi2017minimum}. Furthermore, despite recent advancements in kernel and activation functions across both shallow and deep learning~\citep{parisi2020qrelu}~\citep{parisietayper}~\citep{parisimachine}~\citep{parisi2021hyper}, it is still challenging to derive a kernel function that can aid feature extraction methods to generalise in presence of non-Gaussian and entropic data~\citep{parisi2018decision}~\citep{parisi2020novel}~\citep{parisi2020evolutionary}.
\\

To address these challenges, over the past two decades, there has been an increasing trend in the Artificial Intelligence (AI) community to generate deep models that incorporate feature extraction within their architectures, starting from the Convolutional Neural Network (CNN)~\citep{lecun1995convolutional} to deep CNN-like variations~\citep{krizhevsky2012imagenet}~\citep{simonyan2014very}~\citep{he2016deep}, and, more recently, even deeper models, such as Transformer-type of neural networks ~\citep{vaswani2017attention}. Nevertheless, this school of thought attempts to handle entropy in the data intrinsically, by training deep models that are capable of automated feature extraction on expensive hardware, e.g., Graphics and Tensor Processing Units. 
\\

This trendy reliance on deep models and costly hardware has transformed AI into an elitist field, albeit with a very few exceptions~\citep{chen2019slide}~\citep{liu2021sparse}, as it demands expensive computing resources for predictive analytics. Furthermore, it has jeopardised the explainability of classifications~\citep{gaur2021semantics}, as features are extracted and transformed within deep model architectures, which, \emph{per se}, are already not straightforward to describe to end users. In fact, these limitations have hindered a wide and transparent application of deep learning models in regulated industries, such as finance and healthcare~\citep{samek2019towards}.
\\

To contribute to the democratisation of AI~\citep{garvey2018framework}~\citep{kobayashi2019will}~\citep{ahmed2020democratization}, thus enabling resource-constrained environments to leverage extrinsic and explainable feature extraction methods for AI to have a wider positive impact in society, in this study, a novel kernel-based dimensionality reduction method is proposed to reduce the uncertainty in the data conveyed to improve classification performance. Written in Python and made freely available in 'scikit-learn'~\citep{scikit-learn} for the 'FastICA' class, the proposed feature extraction method is presented as a more reliable approach than the gold standard FastICA method. By virtue of its liberal license, this kernel-based dimensionality reduction method is broadly disseminated within the free software Python library 'scikit-learn'~\citep{scikit-learn}, and it can be used for both academic research and commercial purposes.\\

\newpage


\vskip 0.3in

\section{Methods}

\vskip 0.15in

\subsection{Datasets leveraged from scikit-learn and the UCI ML repository}

\vskip 0.15in

In this study, the following datasets from \emph{The University California Irvine (UCI) ML repository} were used:

\begin{itemize}

\item \href{https://archive.ics.uci.edu/ml/machine-learning-databases/parkinsons/parkinsons.data}{‘Parkinson’s’ dataset}~\citep{little2007exploiting}, which has 23 features corresponding to 195 biomedical voice measurements from 31 people, 23 with Parkinson's disease (PD), to help in detecting PD from speech signals.

\item \href{https://archive.ics.uci.edu/ml/machine-learning-databases/haberman/haberman.data}{‘Haberman survival’ dataset}~\citep{Lim:1999}, with three features (age of patient at time of operation; patient's year of operation; number of positive axillary nodes detected) to predict whether 306 patients who had undergone surgery for breast cancer would have died within 5 years of follow up or survived for longer.

\item \href{https://archive.ics.uci.edu/ml/machine-learning-databases/00519/heart_failure_clinical_records_dataset.csv}{‘Heart failure clinical records’ dataset}~\citep{chicco2020machine}, to predict whether a patient was deceased during the follow-up period, based on 13 clinical features from medical records of 299 patients who had heart failure.

\item ‘SPECTF’ dataset~\citep{Cios:2001}, which has 267 images collected via a cardiac Single Proton Emission Computed Tomography (SPECT), describing whether each patient has a physiological or pathophysiological heart based on 44 features, as per the following two data splits or partitions:

\begin{itemize}
\item \href{https://archive.ics.uci.edu/ml/machine-learning-databases/spect/SPECTF.train}{training data partition} ('SPECTF.train' file), with 80 images.
\item \href{https://archive.ics.uci.edu/ml/machine-learning-databases/spect/SPECTF.test}{testing data partition} ('SPECTF.test' file), which has 187 images.
\end{itemize}

\end{itemize}

Moreover, the following dataset from \emph{scikit-learn} was leveraged for a further evaluation:

\begin{itemize}

\item 'Breast cancer Wisconsin (diagnostic)' dataset~\citep{Wolberg:1995}, having 30 characteristics of cell \emph{nuclei} from 569 digitised images of a fine needle aspirate of breast masses, to detect whether they correspond to either malignant or benign breast cancer.

\end{itemize}

\vskip 0.15in

\subsection{Baseline MLP model, activation functions, and hyperparameters}

\vskip 0.15in

The purpose of this study is not to devise the most optimised, best-performing classifier for any of the classification tasks involved in section 2.1, but to develop a novel reliable kernel-based feature extraction method. Thus, whilst applying early stopping to avoid overfitting, a baseline Multi-Layer Perceptron (MLP)~\citep{rumelhart1986learning}~\citep{parisi2014neural}~\citep{parisi2015novel}~\citep{parisi2018feature}~\citep{parisi2019machine} model was leveraged using various benchmark activation functions (m-arcsinh~\citep{parisi2020m}~\citep{parisim}, identity, tanh~\citep{jacot2018neural}, and ReLU) and with the following hyperparameters for all classification tasks in section 2.1:

\begin{itemize}

\item MLP-related hyperparameters:
\begin{itemize}
\item \emph{'random\_state'} = 1.
\item \emph{'max\_iter'} = 250, where 'max\_iter' is the maximum number of iterations.
\end{itemize}

\end{itemize}

Listing 1 provides the snippet of code in Python to use an MLP with different activation functions available in 'scikit-learn'~\citep{scikit-learn}, including the 'm-arcsinh', following feature extraction.

\lstset{language=Python}
\lstset{frame=lines}
\lstset{caption={MLP with different activation functions available in 'scikit-learn'~\citep{scikit-learn}, including the 'm-arcsinh'.}}
\lstset{label={lst:code_direct_1}}
\lstset{basicstyle=\footnotesize}
\begin{lstlisting}

from sklearn.neural_network import MLPClassifier

for activation in ('m-arcsinh', 'identity', 'tanh', 'relu'):
    classifier =  MLPClassifier(activation=activation, 
    random_state=1, max_iter=1000)

\end{lstlisting}

Except for the ‘SPECTF’ dataset~\citep{Cios:2001}, which is already provided in two separate partitions for training and testing (see section 2.1), the other four datasets were split into 80\% for training and 20\% for testing via \emph{'train\_test\_split'} in 'scikit-learn'~\citep{scikit-learn} from \emph{'sklearn.model\_selection'}, without randomisation (\emph{'shuffle'=False}) to enable reproducibility of the results presented in this study.

\vskip 0.15in

\subsection{m-arK-FastICA: A novel reliable feature extraction approach}

Coupling FastICA with the modified ('m-') inverse hyperbolic sine function ('arcsinh') or 'm-arcsinh' function that was found to generalise as a kernel and activation for optimal separating hyperplane- and neural network-based classifiers~\citep{parisi2020m}~\citep{parisim}, the proposed feature extraction method seeks to 1) handle the non-linearity in the data further via an improved transformation in the G function that is leveraged in the approximation to neg-entropy, 2) cope with entropic data by coupling the m-arcsinh Kernel ('m-ar-K') with FastICA, thus achieving the hybrid method 'm-ar-K-FastICA', and 3) improve classification of non-linearly separable and entropic input data into target classes by fulfilling points 1) and 2).

Via a weighted interaction effect between the arcsinh and the slightly non-linear characteristic of the squared root function, the m-ar-K~\citep{parisi2020m}~\citep{parisim} can be represented as follows:

\vspace{1em}
\begin{math}
y = arcsinh(x) \times \frac{1}{3} \times \frac{1}{4} \times \sqrt{\left|x\right|} = arcsinh(x) \times \frac{1}{12} \times  \sqrt{\left|x\right|}\hspace{13.5em}(1)
\end{math}
\vspace{2em}

\begin{tikzpicture}
\begin{axis}[ymin=-0.1, ymax=0.1, enlargelimits=false]
\addplot [domain=-1:1, samples=101,unbounded coords=jump]{(arcsinh(x)/3)*(sqrt(abs(x))/4)};
\end{axis}
\end{tikzpicture}

\vspace{2em}

The derivative of m-ar-K~\citep{parisi2020m}~\citep{parisim} is described as follows:

\vspace{1em}
\begin{math}
{\frac{dy}{dx} = \sqrt{\left|x\right|}} \times \frac{1}{12\times\sqrt{x^2+1}}  +\  \frac{x\ \times arcsinh(x)}{24\times\left|x\right|^\frac{3}{2}\ }\hspace{21.5em}(2)
\end{math}
\hspace{36.5em}(2)
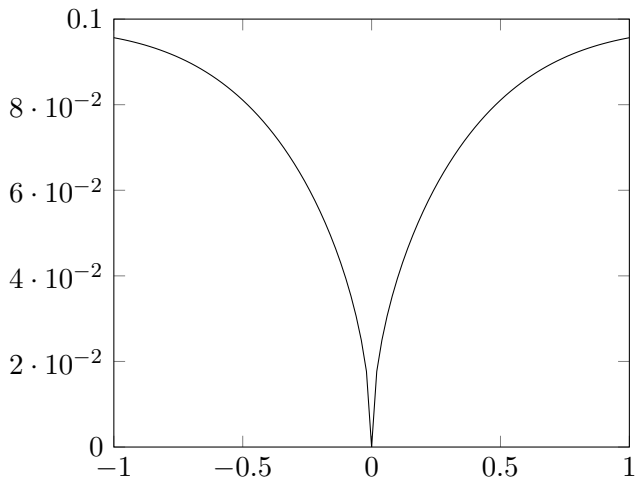
\vspace{2em}

\begin{tikzpicture}
\begin{axis}[ymin=0, ymax=0.1, enlargelimits=false]
\addplot [domain=-1:1, samples=101,unbounded coords=jump]{sqrt(abs(x))/(12*sqrt(x^2+1)) + (x*arcsinh(x))/(24*abs(x)^(3/2)))};
\end{axis}
\end{tikzpicture}

\vspace{2em}

To enable reproducibility of the results obtained in this study, considering that the number of components (~\emph{n\_components}) were set based on the dimensionality and the level of entropy of the data, the following parameters were used in both the FastICA and the proposed m-ar-K-FastICA method to extract features from:

\begin{itemize}
\item All datasets:
\begin{itemize}
\item \emph{'random\_state'} = 42.
\end{itemize}
\end{itemize}

\begin{itemize}
\item The ‘Haberman survival’~\citep{Lim:1999} and the ‘Heart failure clinical records’ datasets~\citep{chicco2020machine}:
\begin{itemize}
\item \emph{'n\_components'} = 2.
\end{itemize}
\end{itemize}

\begin{itemize}
\item The ‘Parkinson's’ dataset~\citep{little2007exploiting}:
\begin{itemize}
\item \emph{'n\_components'} = 8.
\end{itemize}
\end{itemize}

\begin{itemize}
\item The ‘Breast cancer Wisconsin (diagnostic)’ dataset~\citep{Wolberg:1995}:
\begin{itemize}
\item \emph{'n\_components'} = 16.
\end{itemize}
\end{itemize}

\begin{itemize}
\item The ‘SPECTF’ dataset~\citep{Cios:2001}:
\begin{itemize}
\item \emph{'n\_components'} = 43.
\end{itemize}
\end{itemize}

Listing 2 provides the snippet of code in Python that implements the proposed m-arcsinh Kernel (m-ar-K) (1) and its derivative (2) coupled in the 'FastICA' class as a G function leveraged to approximate to neg-entropy in 'scikit-learn'~\citep{scikit-learn}.

\lstset{language=Python}
\lstset{frame=lines}
\lstset{caption={Coupling the m-arcsinh kernel ('m-ar-K') and its derivative as a G function in the approximation to neg-entropy in 'scikit-learn'~\citep{scikit-learn}.}}
\lstset{label={lst:code_direct_4}}
\lstset{basicstyle=\footnotesize}
\begin{lstlisting}

import numpy as np


def _m_arcsinh(x, fun_args):
    """Compute the m-arcsinh kernel function and its derivative 
    in place.
    
    It exploits the fact that the derivative is a simple function of the 
    output value from the m-arcsinh kernel.

    Parameters
    ----------
    x: {array-like, sparse matrix}, shape (n_samples, n_features)
        The input data.

    Returns
    -------
    A tuple containing the value of the function, and that of its derivative.
    """

    return (1/3*np.arcsinh(x))*(1/4*np.sqrt(np.abs(x))),
           (np.sqrt(np.abs(x))/(12*np.sqrt(x**2+1)) 
           + (x*np.arcsinh(x))/(24*np.abs(x)**(3/2))).mean(axis=-1)




# Using the m-arcsinh kernel as a G function ('fun') inside the class 'FastICA'
# for feature extraction (using 16 components and fixed random state set to 42 
# for reproducibility in the example below)

if self.fun == 'logcosh':
    g = _logcosh
elif self.fun == 'exp':
    g = _exp
elif self.fun == 'cube':
    g = _cube
elif self.fun == 'm_arcsinh':
    g = _m_arcsinh
elif callable(self.fun):
    def g(x, fun_args):
        return self.fun(x, **fun_args)
else:
    exc = ValueError if isinstance(self.fun, str) else TypeError
    raise exc(
        "Unknown function %r;"
        " should be one of 'logcosh', 'exp', 'cube', 'm_arcsinh', or callable"
        % self.fun
    )


from sklearn.decomposition import FastICA
transformer = FastICA(n_components=16, random_state=42, fun='m_arcsinh')
transformed_inputs = transformer.fit_transform(inputs)

\end{lstlisting}

\vskip 0.15in

\subsection{Performance evaluation}

\vskip 0.15in

Further to feature extraction as per section 2.3, the accuracy of the MLP classifier in section 2.2 using different activation functions (m-arcsinh~\citep{parisi2020m}~\citep{parisim}, identity, tanh~\citep{jacot2018neural}, and ReLU) on the datasets described in section 2.1, was evaluated via the \emph{'accuracy\_score'} available in 'scikit-learn'~\citep{scikit-learn} from \emph{'sklearn.metrics'}. 
The reliability of such classifiers was assessed via the weighted average of the precision, recall and F1-score computed via the \emph{'classification\_report'}, also available in 'scikit-learn'~\citep{scikit-learn} from \emph{'sklearn.metrics'}.
\\
To understand what classification accuracy and reliability are, and how they can be evaluated, please refer to the following studies:~\citep{parisi2018decision},~\citep{parisi2018feature},~\citep{parisi2020novel},~\citep{parisi2020evolutionary}.

Moreover, the computational cost of the classifier, to quantify the impact of using the proposed feature extraction method (m-ar-K-FastICA) as opposed to the current gold standard one (FastICA), was assessed via the training time in seconds. Experiments were run on an AMD E2-9000 Radeon R2 processor, 1.8 GHz and 4 GB DDR4 RAM.

\newpage


\section{Results}

\vskip 0.15in

Experimental results demonstrate the overall higher reliability achieved via the proposed m-ar-K-FastICA kernel-based dimensionality reduction method for MLP-based classification, including a shorter training time, as follows:
\begin{itemize}
\item Higher reliability on 2 of 5 datasets evaluated (Tables 1 and 3).
\item The same classification performance on 3 of 5 datasets assessed (Tables 2, 4, and 5).
\item Shorter training time for 3 of 5 datasets evaluated (Tables 2-4).
\end{itemize}

\vskip 0.145in


\textbf{Table 1} Results on performance evaluation of baseline (non-optimised) Multi-Layer Perceptron (MLP) in scikit-learn with different kernel functions, to compare the impact of using either the gold standard FastICA feature extraction algorithm or the proposed m-ar-K-FastICA method. The performance of this classifier was assessed on the ‘Breast cancer Wisconsin (diagnostic)’ dataset~\citep{Wolberg:1995} available in scikit-learn.

\begin{table}[H]
\resizebox{\textwidth}{!}{%
\begin{tabular}{llllllll}
\textbf{Classifier} &
  \textbf{Kernel} &
  \textbf{Feature extraction} &
  \textbf{\begin{tabular}[c]{@{}l@{}}Training time \\ \\ (s)\end{tabular}} &
  \textbf{\begin{tabular}[c]{@{}l@{}}Accuracy \\ \\ (0-1)\end{tabular}} &
  \textbf{\begin{tabular}[c]{@{}l@{}}Precision \\ \\ (0-1)\end{tabular}} &
  \textbf{\begin{tabular}[c]{@{}l@{}}Recall \\ \\ (0-1)\end{tabular}} &
  \textbf{\begin{tabular}[c]{@{}l@{}}F1-score \\ \\ (0-1)\end{tabular}} \\
MLP & m-arcsinh & FastICA & 4.71 & 0.95 & 0.95 & 0.95 & 0.95 \\
MLP & Identity & FastICA & 1.64 & 0.95 & 0.95 & 0.95 & 0.95 \\
MLP & tanh & FastICA & 2.97 & 0.97 & 0.97 & 0.97 & 0.97 \\
MLP & ReLU & FastICA & 1.87 & 0.95 & 0.95 & 0.95 & 0.95 \\
MLP & m-arcsinh & \textbf{m-ar-K-FastICA} & 5.47 & 0.95 & 0.95 & 0.95 & 0.95 \\
MLP & Identity & \textbf{m-ar-K-FastICA} & 1.12 & 0.96 & 0.97 & 0.96 & 0.96 \\
MLP & tanh & \textbf{m-ar-K-FastICA} & 3.40 & 0.97 & 0.97 & 0.97 & 0.97 \\
MLP & ReLU & \textbf{m-ar-K-FastICA} & 2.00 & 0.96 & 0.97 & 0.96 & 0.96
\end{tabular}%
}
\end{table}


\textbf{Table 2.} Results on performance evaluation of baseline (non-optimised) Multi-Layer Perceptron (MLP) in scikit-learn with different kernel functions, to compare the impact of using either the gold standard FastICA feature extraction algorithm or the proposed m-ar-K-FastICA method. The performance of this classifier was assessed on the ‘Heart failure clinical records’ dataset~\citep{chicco2020machine} available in the University California Irvine (UCI) Machine Learning repository.

\begin{table}[H]
\resizebox{\textwidth}{!}{%
\begin{tabular}{llllllll}
\textbf{Classifier} &
  \textbf{Kernel} &
  \textbf{Feature extraction} &
  \textbf{\begin{tabular}[c]{@{}l@{}}Training time \\ \\ (s)\end{tabular}} &
  \textbf{\begin{tabular}[c]{@{}l@{}}Accuracy \\ \\ (0-1)\end{tabular}} &
  \textbf{\begin{tabular}[c]{@{}l@{}}Precision \\ \\ (0-1)\end{tabular}} &
  \textbf{\begin{tabular}[c]{@{}l@{}}Recall \\ \\ (0-1)\end{tabular}} &
  \textbf{\begin{tabular}[c]{@{}l@{}}F1-score \\ \\ (0-1)\end{tabular}} \\
MLP & m-arcsinh & FastICA & 0.29 & 0.78 & 0.61 & 0.78 & 0.69 \\
MLP & Identity & FastICA & 0.26 & 0.78 & 0.61 & 0.78 & 0.69 \\
MLP & tanh & FastICA & 0.34 & 0.78 & 0.61 & 0.78 & 0.69 \\
MLP & ReLU & FastICA & 0.30 & 0.78 & 0.61 & 0.78 & 0.69 \\
MLP & m-arcsinh & \textbf{m-ar-K-FastICA} & 0.29 & 0.78 & 0.61 & 0.78 & 0.69 \\
MLP & Identity & \textbf{m-ar-K-FastICA} & 0.47 & 0.78 & 0.61 & 0.78 & 0.69 \\
MLP & tanh & \textbf{m-ar-K-FastICA} & 0.34 & 0.78 & 0.61 & 0.78 & 0.69 \\
MLP & ReLU & \textbf{m-ar-K-FastICA} & 0.38 & 0.78 & 0.61 & 0.78 & 0.69
\end{tabular}%
}
\end{table}

\newpage


\textbf{Table 3.} Results on performance evaluation of baseline (non-optimised) Multi-Layer Perceptron (MLP) in scikit-learn with different kernel functions, to compare the impact of using either the gold standard FastICA feature extraction algorithm or the proposed m-ar-K-FastICA method. The performance of this classifier was assessed on the ‘Haberman survival’ dataset~\citep{Lim:1999} available in the University California Irvine (UCI) Machine Learning repository.

\begin{table}[H]
\resizebox{\textwidth}{!}{%
\begin{tabular}{llllllll}
\textbf{Classifier} &
  \textbf{Kernel} &
  \textbf{Feature extraction} &
  \textbf{\begin{tabular}[c]{@{}l@{}}Training time \\ \\ (s)\end{tabular}} &
  \textbf{\begin{tabular}[c]{@{}l@{}}Accuracy \\ \\ (0-1)\end{tabular}} &
  \textbf{\begin{tabular}[c]{@{}l@{}}Precision \\ \\ (0-1)\end{tabular}} &
  \textbf{\begin{tabular}[c]{@{}l@{}}Recall \\ \\ (0-1)\end{tabular}} &
  \textbf{\begin{tabular}[c]{@{}l@{}}F1-score \\ \\ (0-1)\end{tabular}} \\
MLP & m-arcsinh & FastICA & 0.93 & 0.82 & 0.68 & 0.82 & 0.74 \\
MLP & Identity & FastICA & 0.95 & 0.82 & 0.68 & 0.82 & 0.74 \\
MLP & tanh & FastICA & 0.96 & 0.82 & 0.68 & 0.82 & 0.74 \\
MLP & ReLU & FastICA & 1.04 & 0.82 & 0.68 & 0.82 & 0.74 \\
MLP & m-arcsinh & \textbf{m-ar-K-FastICA} & 0.80 & 0.82 & 0.78 & 0.82 & 0.79 \\
MLP & Identity & \textbf{m-ar-K-FastICA} & 1.10 & 0.82 & 0.78 & 0.82 & 0.79 \\
MLP & tanh & \textbf{m-ar-K-FastICA} & 0.92 & 0.82 & 0.78 & 0.82 & 0.79 \\
MLP & ReLU & \textbf{m-ar-K-FastICA} & 1.00 & 0.82 & 0.78 & 0.82 & 0.79
\end{tabular}%
}
\end{table}


\textbf{Table 4.} Results on performance evaluation of baseline (non-optimised) Multi-Layer Perceptron (MLP) in scikit-learn with different kernel functions, to compare the impact of using either the gold standard FastICA feature extraction algorithm or the proposed m-ar-K-FastICA method. The performance of this classifier was assessed on the ‘Parkinson’s’ dataset~\citep{little2007exploiting} available in the University California Irvine (UCI) Machine Learning repository.

\begin{table}[H]
\resizebox{\textwidth}{!}{%
\begin{tabular}{llllllll}
\textbf{Classifier} &
  \textbf{Kernel} &
  \textbf{Feature extraction} &
  \textbf{\begin{tabular}[c]{@{}l@{}}Training time \\ \\ (s)\end{tabular}} &
  \textbf{\begin{tabular}[c]{@{}l@{}}Accuracy \\ \\ (0-1)\end{tabular}} &
  \textbf{\begin{tabular}[c]{@{}l@{}}Precision \\ \\ (0-1)\end{tabular}} &
  \textbf{\begin{tabular}[c]{@{}l@{}}Recall \\ \\ (0-1)\end{tabular}} &
  \textbf{\begin{tabular}[c]{@{}l@{}}F1-score \\ \\ (0-1)\end{tabular}} \\
MLP & m-arcsinh & FastICA & 0.80 & 0.77 & 0.59 & 0.77 & 0.67 \\
MLP & Identity & FastICA & 0.78 & 0.77 & 0.59 & 0.77 & 0.67 \\
MLP & tanh & FastICA & 0.80 & 0.77 & 0.59 & 0.77 & 0.67 \\
MLP & ReLU & FastICA & 0.68 & 0.77 & 0.59 & 0.77 & 0.67 \\
MLP & m-arcsinh & \textbf{m-ar-K-FastICA} & 0.67 & 0.77 & 0.59 & 0.77 & 0.67 \\
MLP & Identity & \textbf{m-ar-K-FastICA} & 0.62 & 0.77 & 0.59 & 0.77 & 0.67 \\
MLP & tanh & \textbf{m-ar-K-FastICA} & 0.63 & 0.77 & 0.59 & 0.77 & 0.67 \\
MLP & ReLU & \textbf{m-ar-K-FastICA} & 0.70 & 0.77 & 0.59 & 0.77 & 0.67
\end{tabular}%
}
\end{table}

\newpage


\textbf{Table 5.} Results on performance evaluation of baseline (non-optimised) Multi-Layer Perceptron (MLP) in scikit-learn with different kernel functions, to compare the impact of using either the gold standard FastICA feature extraction algorithm or the proposed m-ar-K-FastICA method. The performance of this classifier was assessed on the ‘SPECTF’ dataset~\citep{Cios:2001} available in the University California Irvine (UCI) Machine Learning repository.

\begin{table}[H]
\resizebox{\textwidth}{!}{%
\begin{tabular}{llllllll}
\textbf{Classifier} &
  \textbf{Kernel} &
  \textbf{Feature extraction} &
  \textbf{\begin{tabular}[c]{@{}l@{}}Training time \\ \\ (s)\end{tabular}} &
  \textbf{\begin{tabular}[c]{@{}l@{}}Accuracy \\ \\ (0-1)\end{tabular}} &
  \textbf{\begin{tabular}[c]{@{}l@{}}Precision \\ \\ (0-1)\end{tabular}} &
  \textbf{\begin{tabular}[c]{@{}l@{}}Recall \\ \\ (0-1)\end{tabular}} &
  \textbf{\begin{tabular}[c]{@{}l@{}}F1-score \\ \\ (0-1)\end{tabular}} \\
MLP & m-arcsinh & FastICA & 0.53 & 1.00 & 1.00 & 1.00 & 1.00 \\
MLP & Identity & FastICA & 0.60 & 1.00 & 1.00 & 1.00 & 1.00 \\
MLP & tanh & FastICA & 0.62 & 1.00 & 1.00 & 1.00 & 1.00 \\
MLP & ReLU & FastICA & 0.65 & 1.00 & 1.00 & 1.00 & 1.00 \\
MLP & m-arcsinh & \textbf{m-ar-K-FastICA} & 0.68 & 1.00 & 1.00 & 1.00 & 1.00 \\
MLP & Identity & \textbf{m-ar-K-FastICA} & 0.53 & 1.00 & 1.00 & 1.00 & 1.00 \\
MLP & tanh & \textbf{m-ar-K-FastICA} & 0.65 & 1.00 & 1.00 & 1.00 & 1.00 \\
MLP & ReLU & \textbf{m-ar-K-FastICA} & 0.81 & 1.00 & 1.00 & 1.00 & 1.00
\end{tabular}%
}
\end{table}

\newpage


\vskip 0.3in

\section{Discussion}

\vskip 0.15in

As demonstrated by the highest reliability achieved via the proposed feature extraction method 'm-ar-K-FastICA', quantified in section 2.4, and achieved on 2 datasets assessed (Tables 1 and 3), whilst the same reliability was obtained on the other 3 datasets evaluated, it yielded the best classification performance when considering the same classifier as opposed to the current gold standard FastICA method. For the dataset considered in Table 1, when using the MLP with the identity and ReLU activation functions, leveraging the novel m-ar-K-FastICA method led to 1\% improvement in classification accuracy, recall/sensitivity, and F1-score, whilst the precision increased by 2\%. When considering the dataset in Table 3, precision and F1-score increased by 10\% and 5\% respectively, using the MLP classifier for all activation functions used (m-arcsinh, identity, tanh, ReLU).
\\

Thus, overall, the m-ar-K-FastICA algorithm led to less false positives and increased reliability of the predictions. These improvements in classification performance achieved via the proposed dimensionality reduction method is important to consider for medical applications to reduce psychosocial consequences due to inaccurate screenings~\citep{rasmussen2020psychosocial}, such as those to aid early diagnosis of various tumours, including lung, prostate, and breast cancers.
\\

Besides its higher reliability, leveraging the proposed m-ar-K-FastICA feature extraction method led to reductions in training time for 3 of 5 datasets evaluated, i.e., 14.38\%, 3.27\%, and 1.55\% shorter training times for datasets in Tables 4, 2, and 3 respectively. Thus, also considering the computational efficiency brought by the novel dimensionality reduction method m-ar-K-FastICA, it shows potential to be applied in resource-constrained environments for maximising its positive impact.
\\

Therefore, the m-ar-K-FastICA demonstrates that the m-arcsinh kernel is not only a generalisable function for classification, but also when coupled with feature extraction methods, such as FastICA, to improve the reliability and computational efficiency of classifiers.
As a reliable and computationally efficient approach, the m-ar-K-FastICA is thus deemed a new gold standard dimensionality reduction method, freely available in scikit-learn. 

\newpage


\section{Conclusion}

The m-ar-K-FastICA feature extraction algorithm in scikit-learn provides a new method to achieve dimensionality reduction, even in presence of entropic data. In this study, the proposed approach was shown to lead to higher reliability and computationally efficiency in supervised learning, with regards to MLP-aided classification. As it is made freely available, open source, on the Python and scikit-learn ecosystems, it adds to the choices that both academia and industry can have when selecting a dimensionality reduction method, also in resource-constrained environments. 
\\

Importantly, the proposed m-ar-K-FastICA method was found faster and more reliable for feature extraction than the current gold standard FastICA method. Written in a high-level programming language (Python), it can be used within ML-based pipelines in specific use cases, wherein high precision and reliability need to be obtained, whilst expensive computational hardware may not be available, such as to aid screenings of tumours, including prostate, lung, and breast cancers. Future work involves extending the coupling of the m-arcsinh kernel to benefit other dimensionality reduction methods and provide further choices that the AI community worldwide, including in resource-constrained environments, can have when selecting an extrinsic feature extraction algorithm to aid predictive analytics.


\acks{This research did not receive any specific grant from funding agencies in the public, commercial, or not-for-profit sectors.}


\newpage

\bibliography{ArXiv_paper_LP}

\end{document}